\documentclass[final,5p,times,twocolumn]{elsarticle}

\usepackage{lineno,hyperref}
\modulolinenumbers[5]

\journal{}

\usepackage{amsmath}
\usepackage{color}
\usepackage{subfigure,amsfonts,amssymb,booktabs,hyperref}
\usepackage{stfloats}
\usepackage{enumerate}
\usepackage{color}

\biboptions{numbers,sort&compress}
\def\etal{{\emph{et al. }}}

\usepackage{xparse}%

\usepackage{blindtext}

\usepackage{morewrites}

\usepackage{wrapfig}
\usepackage{xkeyval}%

\newwrite\authorbibfile%

\AtBeginDocument{%
  \immediate\openout\authorbibfile=\jobname.aub%
}%

\AtEndDocument{%
\immediate\closeout\authorbibfile
\InputIfFileExists{\jobname.aub}{}{}
}%

\makeatletter

\define@key{authorbib}{scale}[1]{%
\def\AuthorbibKVMacroScale{#1}%
}

\define@key{authorbib}{wraplines}[10]{%
\def\AuthorbibKVMacroWraplines{#1}%
}

\define@key{authorbib}{imagewidth}[4cm]{%
\def\AuthorbibKVMacroImagewidth{#1}%
}

\define@key{authorbib}{overhang}[10pt]{%
\def\AuthorbibKVMacroOverhang{#1}%
}

\define@key{authorbib}{imagepos}[l]{%
\def\AuthorbibKVMacroImagepos{#1}%
}

\renewcommand{\@thesubfigure}{\hskip\subfiglabelskip}
\makeatother
\presetkeys{authorbib}{imagepos=l,imagewidth=4cm,wraplines=15,overhang=20pt}{}

\newlength{\AuthorbibTopSkip}
\newlength{\AuthorbibBottomSkip}
\NewDocumentCommand{\authorbibliography}{+o+m+m+m}{%
  \IfNoValueTF{#1}{%
  }{%
    \setkeys{authorbib}{#1}%
    \immediate\write\authorbibfile{%
      \string\begin{wrapfigure}[\AuthorbibKVMacroWraplines]{\AuthorbibKVMacroImagepos}[\AuthorbibKVMacroOverhang]{\AuthorbibKVMacroImagewidth}^^J
        \string\includegraphics[scale=\AuthorbibKVMacroScale]{#2}^^J
        \string\end{wrapfigure}^^J
    }%
  }%
  \IfNoValueTF{#3}{%
    \typeout{Warning: No author name}%
  }{%
    \immediate\write\authorbibfile{%
      \unexpanded{\vspace{\AuthorbibTopSkip}}^^J
      \string\noindent\relax
      \unexpanded{\textbf{#3}\par}^^J
      \string\noindent\relax
      \unexpanded{#4}^^J%
      \unexpanded{\vspace{\AuthorbibBottomSkip}}^^J
      }%
  }%
}%

\begin{document}

\begin{frontmatter}

\title{Data augmentation by morphological mixup for solving Raven's Progressive Matrices}

\author{Wentao He}

\author{Jianfeng Ren\corref{mycorrespondingauthor}}
\ead{jianfeng.ren@nottingham.edu.cn}

\author{Ruibin Bai}

\cortext[mycorrespondingauthor]{Corresponding author. Tel.: +86 (0)574 8818 0000--8805}

\address{The School of Computer Science, University of Nottingham Ningbo China, \\199 Taikang East Road, Ningbo, 315100 China}

\begin{abstract}
Raven’s Progressive Matrices (RPMs) are frequently used in testing human’s visual reasoning ability. Recent advances of RPM-like datasets and solution models partially address the challenges of visually understanding the RPM questions and logically reasoning the missing answers. In view of the poor generalization performance due to insufficient samples in RPM datasets, we propose an effective scheme, namely Candidate Answer Morphological Mixup (CAM-Mix). CAM-Mix serves as a data augmentation strategy by gray-scale image morphological mixup, which regularizes various solution methods and overcomes the model overfitting problem. By creating new negative candidate answers semantically similar to the correct answers, a more accurate decision boundary could be defined. By applying the proposed data augmentation method, a significant and consistent performance improvement is achieved on various RPM-like datasets compared with the state-of-the-art models.
\end{abstract}


\end{frontmatter}


\section{Introduction}
\label{sec:intro}
Visual recognition has been widely and deeply explored, while visual reasoning \cite{xiao2019daa,antol2015vqa,chu2018forgettable,hong2019exploiting,gao2020question,lao2021multi,zhang2019raven,hu2021stratified} recently becomes an increasingly important research topic and needs further investigation. 
Visual reasoning often consists of two tasks: visual recognition and logical reasoning. It leverages on the development of visual recognition while pushing beyond it. Visual reasoning infers the relationship between objects based on the perceived visual attributes. In computer vision community, data augmentation techniques have been proved effective for many problems, including the latest developments in image classification \cite{yun2019cutmix,wang2021regularizing}, object detection \cite{chen2020self,xia2021visible}, synthetic data generation \cite{gatys2015neural,zhu2017unpaired} and many others \cite{leng2017data,fu2020multi,du2020object,liu2020novel,feng2020autuencoder}. In this paper, we aim to improve the performance of visual reasoning system through data augmentation, especially for the problem of solving visual IQ tests \cite{zhang2019raven,hu2021stratified}.

\begin{figure}[htbp]
    \centering
    \subfigure[(a) Question]{
        \begin{minipage}[t]{0.45\linewidth}
            \centering
            \includegraphics[width=4cm]{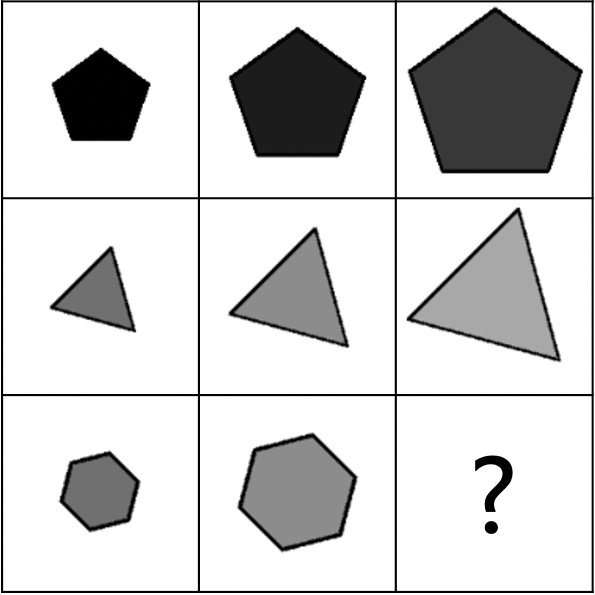}
        \end{minipage}
    }
    \subfigure[(b) Candidate answers]{
        \begin{minipage}[t]{0.45\linewidth}
            \centering
            \includegraphics[width=4cm]{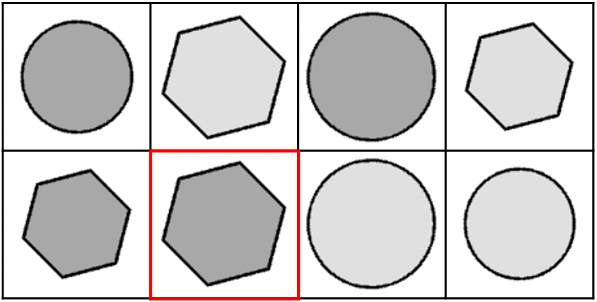}
        \end{minipage}
    }
    \centering
    \caption{An example of the I-RAVEN dataset \cite{hu2021stratified} with question panel (a) and candidate answer set (b). The correct answer is marked in red. }
    \label{example}
\end{figure}

Raven's Progressive Matrix (RPM) \cite{raven1938raven,kunda2013computational} problem is one of the frequently-used tests on human visual reasoning in cognitive science. An RPM problem is formed by a 3$\times$3 pictorial matrix with the last one left blank. The task is to identify the missing entry from 8 candidate answers. One illustrative example is shown in Fig.~\ref{example}. As the RPM test assesses human's visual reasoning ability using pictorial matrices containing visually simple patterns, it minimizes the impact of language barrier and culture bias \cite{raven1938raven}. 

Researchers have made considerable efforts to develop a system to automatically solve the RPM problem as part of broad research initiatives for advanced automated visual reasoning \cite{zhang2019raven,santoro2018measuring,hu2021stratified,zhang2019learning,zheng2019abstract,kim2020few}. Santoro \etal \cite{santoro2018measuring} constructed the first RPM-like dataset named Procedurally Generated Matrices and developed a reasoning algorithm named Wild Relation Network (WReN). Zhang \etal \cite{zhang2019raven} developed the dataset called Relational and Analogical Visual rEasoNing (RAVEN), and a ResNet architecture assembled with a Dynamic Residual Tree (DRT) module for reasoning. Later Zhang \etal \cite{zhang2019learning} developed the CoPINet architecture under the principle of contrasting, achieving the state-of-the-art performance on the RAVEN dataset. Very recently, Hu \etal \cite{hu2021stratified} argued that the RAVEN dataset \cite{zhang2019raven} has a shortcut to solutions, and developed an Impartial-RAVEN dataset and a solution model named Stratified Rule-Aware Network (SRAN). 

In RPM problems such as RAVEN, besides the correct answer, 7 incorrect answers often serve as the negative candidate answers to confuse humans. 
From the machine-learning point of view, these negative candidate answers will result in an increasing number of training samples, which is beneficial to the model training process. More specifically, more negative candidate answers will define the decision boundary more accurately. 
Thus, we develop a data augmentation algorithm by generating more negative candidate answers, train the models with more samples and consequently improve the model performance. 

The common image manipulations in data augmentation such as image rotation, cropping and flipping are beneficial to visual recognition of natural images \cite{krizhevsky2012imagenet}. However, these approaches are not suitable for RPM problems, as they only complicate the visual recognition process, but do not help the reasoning process. For example, the visual recognition system needs additional efforts to identify the image rotation angle, while it is an irrelevant attribute in RPM problems, which does not help the reasoning process and may even confuse the logical reasoning system.

In this paper, 
we propose a data augmentation method named \textit{{C}andidate {A}nswer {M}orphological {Mix}up} ({CAM-Mix}), a gray-level pixel-wise morphological operation that convexly combines candidate answers, by mixing up the correct answer candidate with negative candidates. The generated negative samples are semantically similar but different from the correct answers, and hence could define a more accurate boundary around the space spanned by the correct answers. In addition, many visual attributes remain unchanged in the proposed CAM-Mix, which avoids the problems of introducing additional attribute values by the traditional image mixup \cite{zhang2018mixup}. 
To the best of our knowledge, this is the first research attempt of introducing data augmentation in solving the RPM problems. 


The main contributions of this paper are two-fold: 
1) We propose a data augmentation mechanism by mixing up the candidate answers, which could define a more accurate decision boundary around the space spanned by the correct answers. 2) As a plug-and-play tool, the proposed method achieves a significant and consistent improvement compared with the state-of-the-art models on RPM problems. 

\section{Related work}
\label{sec:relatedwork}
\subsection{Data augmentation}
With great advances in recently developed deep convolutional neural networks, remarkable performance has been achieved for various computer vision tasks \cite{ren2013complete,he2016deep,badrinarayanan2017segnet,ren2017regularized,wang2019blood,ren2021three}. Sufficient training data often serve as the foundation to guarantee the performance of deep learning models \cite{leng2017data}. To improve the generalization ability of the models, data augmentation has been widely used as an effective scheme to assist the model training process. According to the survey by Shorten \etal \cite{shorten2019survey}, image data augmentation techniques can be categorized into two main types, basic image manipulations and deep learning approaches. 

In the first category, geometric transformations like image rotation, flipping and cropping perform well on natural images \cite{krizhevsky2012imagenet}, but they are generally task-dependent and may not be safely applied in specific shift-variant tasks, e.g. applying image rotation in digit recognition tasks of 6 verses 9. Later, the development of mixing images and random erasing \cite{zhong2020random} guarantees better generalization ability on various tasks. Specifically, the idea of mixing images can be classified into non-linearly mixup, i.e. assembling of sample instances \cite{inoue2018data,summers2019improved} or sub-images \cite{takahashi2019data}, and a linear interpolation of image pairs such as alpha-blending \cite{zhang2018mixup}. 
Vanilla mixup \cite{zhang2018mixup} uses a simple linear interpolation function which may suffer from overfitting to corrupted samples. To overcome this challenge, researchers have developed mixup variations including Manifold Mixup \cite{verma2019manifold}, which leverages semantic interpolations inside layers to train neural networks, and MetaMixUp \cite{mai2021metamixup}, which utilizes the concept of meta-learning to determine an optimized interpolation policy to regularize network models. 

Data augmentation based on deep learning usually refers to adversarial approaches, such as transfer learning \cite{liu2018feature} or Generative Adversarial Network (GAN) \cite{fu2020multi,du2020object,liu2020novel}. They can generate more synthetic data visually similar to existing training data and help to improve the classification performance. Such data augmentation techniques related to GAN models are usually more task-specific, and hence the generalization ability across different tasks may be limited. Fu \etal \cite{fu2020multi} developed a GAN-based data augmentation model for fine-grained classification problems, which can produce class-dependent synthetic images with fine-grained details. GAN-based data augmentation adopted in \cite{du2020object} aims to handle the problem of sample inadequacy and class imbalance, and consequently facilitates the training of LSTM network in real-time visual tracking problems. Liu \etal \cite{liu2020novel} developed a GAN model for pedestrian detection transferring pedestrians from other datasets into the target scene, to cope with the challenge of insufficient data and heavy pedestrian occlusions. 

\subsection{Visual reasoning}
In literature, visual reasoning spans a variety of tasks, e.g. action recognition, image/video captioning \cite{xiao2019daa}, visual question answering \cite{chu2018forgettable,hong2019exploiting,antol2015vqa,gao2020question,lao2021multi}, and visual IQ tests \cite{zhang2019raven,hu2021stratified}. The first two focus more on visual recognition whereas the last two focus more on relational reasoning.

Visual question answering (VQA) is a conventional visual reasoning task that measures the machine understanding of scene-level images. The objective is to derive an accurate natural language answer, given an image and a related natural language question \cite{antol2015vqa}. 
Recent advances on various solution models lead to a better machine reasoning capability. Hong \etal \cite{hong2019exploiting} developed a neural model to exploit intermediate CNN layers and low-level layer features to answer low-level semantic questions. Gao \etal \cite{gao2020question} further explored the relationship between question semantics and fine-grained object information, which improves the VQA performance. 

Raven's Progressive Matrices (RPMs) are originally designed as a non-verbal assessment for human intelligence. By using pictorial matrices containing visually simple patterns, it minimizes the impact of language barrier and culture bias. Recently, large-scale RPM-style datasets, RAVEN \cite{zhang2019raven} and its variants \cite{hu2021stratified}, were developed to extend the RPM-related study from cognitive science to computer science. The RPM-style datasets \cite{zhang2019raven,hu2021stratified} are often automatically generated, and the problems are solved with minimal prior knowledge about the internal construction rules. 

Santoro \etal \cite{santoro2018measuring} developed the Wild Relation Network (WReN), which applies a Relation Network to model the relationship between the question panel and the candidate answer. Zhang \etal \cite{zhang2019raven} developed a Dynamic Residual Tree (DRT) module that can help logical reasoning in neural networks \cite{zhang2019raven}. They later developed an improved network architecture called CoPINet\cite{zhang2019learning} under the principle of contrasting. Hu \etal \cite{hu2021stratified} argued that RAVEN has a shortcut to solutions, and hence developed an I-RAVEN dataset balancing the candidate answer sets. Their developed solution model, SRAN \cite{hu2021stratified}, achieves the state-of-the-art results on the I-RAVEN dataset.

Various models have been developed in literature to solve the RPM-like problems \cite{santoro2018measuring,zhang2019raven,zheng2019abstract,zhang2019learning,hu2021stratified}. However, to the best of our knowledge, no work has considered the problem of insufficient data for precisely conducting relational reasoning in RPMs. In this paper, we propose a data augmentation method that can effectively improve the visual reasoning capabilities of various state-of-the-art solution models. 

\section{Proposed CAM-Mix by image morphological operation}
\label{sec:method}
\subsection{Problem formulation}
An example of RPM-like problems is shown in Fig.~\ref{example}. 
Given a 3$\times$3 RPM-like question matrix with the last one missing $\boldsymbol{Q}=\{q_0,q_1,\cdots, q_7\}$, the aim is to find the missing image $a^*$ from the candidate answer set $\boldsymbol{A}=\{a_0,a_1,\cdots, a_7\}$. 

We follow the formulation widely adopted in recently developed solution methods based on deep neural models \cite{santoro2018measuring,zhang2019learning,hu2021stratified,zheng2019abstract}, i.e. by training a customized model containing functional reasoning modules using sufficient labelled training data. The model is able to generalize on testing data with input $\boldsymbol{Q}$ and each of candidates $a_i \in \boldsymbol{A}$, and output one predicted answer ${a}^*$ from candidates. The underlying principle for aforementioned algorithms is to make a selection from the candidate answers with maximum probability of being the correct answer given $\boldsymbol{Q}$, $\boldsymbol{A}$: 
\begin{align}\label{eq:1}
    a^* = \arg\max_{a_i \in \boldsymbol{A}} P(a_i|\boldsymbol{Q},\boldsymbol{A}).
\end{align}

The probability of each candidate $P(a_i|\boldsymbol{Q},\boldsymbol{A})$ is derived through the deep convolutional neural network. The model takes a group of 8 images $\boldsymbol{Q}$ and one candidates $a_i$ from $\boldsymbol{A}$ as the input. In CoPINet \cite{zhang2019learning}, for instance, where the model is built upon the Perception Branch $\mathcal{F}$, the Contrast Module $\mathcal{C}$, and the Multilayer Perceptron $\mathcal{M}$, the probability scores $s_i$ for candidate answers are calculated as: 
\begin{align}\label{eq:2}
    \forall a_i \in \boldsymbol{A} \quad s_i & = \mathcal{C}\left( P(a_i|\boldsymbol{Q},\boldsymbol{A})\right) \nonumber \\ 
    & = \mathcal{C}\left( \mathcal{M}\left( \mathcal{F}(\boldsymbol{Q} \cup a_i) \right) \right).
\end{align}



\subsection{Proposed CAM-Mix by gray-level morphological operation}
The idea of mixing up images, which is counter-intuitive to data augmentation \cite{shorten2019survey}, has proven its effectiveness by a line of works \cite{inoue2018data,summers2019improved,liang2018understanding,verma2019manifold,guo2019mixup,mai2021metamixup}. In the traditional vanilla image mixup, the following linear interpolations \cite{zhang2018mixup} are applied on pairs of samples $x_i$, $x_j$ and their one-hot labels $y_i$, $y_j$ in training: 
\begin{align}\label{eq:3}
    &\tilde{x}=\lambda x_i+(1-\lambda)x_j, \nonumber \\
    &\tilde{y}=\lambda y_i+(1-\lambda)y_j. 
\end{align}

Visually, the hyper-parameter $\lambda$ determines how two images $x_i$ and $x_j$ are mixed. From human's perspective, if $\lambda \approx 0.5$ the two images are more likely to be displayed in balanced transparency. The mixed-up images may confuse humans when making decisions, as the generated synthetic images are visually messy and chaotic from human's perspective. However, it may help machine learning models to learn some hidden  relationships in undiscovered search space, and consequently gain a stable performance improvement. 


The RPM-like problems are distinct from general image recognition tasks in the following aspects, which may limit the performance of traditional mixup defined in Eqn. (\ref{eq:3}). 1) In RPM problems, the emphasis is on logical reasoning, and hence the input images are formed up by visually simple patterns, e.g. combinations of regular polygons in a gray-scale intensity form. The image mixup should help more in logic reasoning rather than its original objective in visual recognition. 2) The visual attributes in RPM-like problems come from a finite set. The vanilla mixup defined in Eqn. (\ref{eq:3}) alpha-blends a pair of image samples, which will produce new visual attribute values such as image intensities outside the finite set. These newly introduced values may lead to incorrect reasoning rules and unnecessarily complicate the visual reasoning process. 3) Finally, the vanilla mixup method interpolates the image content as well as the associated labels, which is contradictory to the nature of visual reasoning on RPMs that each question sample can only have exactly one correct answer.

To address these challenges, we propose a mixup algorithm significantly different from previous ones, based on the concept of gray-level image morphological operations \cite{maragos1989representation}. The proposed image mixup method does not generate images with new visual attribute values outside the finite set, but mixes up the visual attributes of two images, without explicitly extracting the visual attributes. Formally, we define the following gray-level morphological operations for a pair of gray-scale images $I_1$ and $I_2$ of the size $n$$\times$$m$: 
\begin{align}
    & \mathcal{OR}(I_1,I_2) = \min_{i \in n, j \in m} \left\{I_1(i,j),I_2(i,j)\right\}, \label{eq:4} \\
    & \mathcal{AND}(I_1,I_2) = \max_{i \in n, j \in m} \left\{I_1(i,j),I_2(i,j)\right\}. \label{eq:5}
\end{align}

One example of the proposed image mixup operation is shown in Fig.~\ref{example_mixup}. We can see that no new visual attribute values (intensity values, shape variations, etc.) are generated during this process. The new images are generated without additional scalars or interpolation functions like vanilla mixup, and the attributes of the generated images are from one of the two original images, without explicitly extracting the attributes. 

\subsection{Data augmentation by proposed CAM-Mix}
\begin{figure*}[pt]
    \centering
    \includegraphics[width=0.6\linewidth]{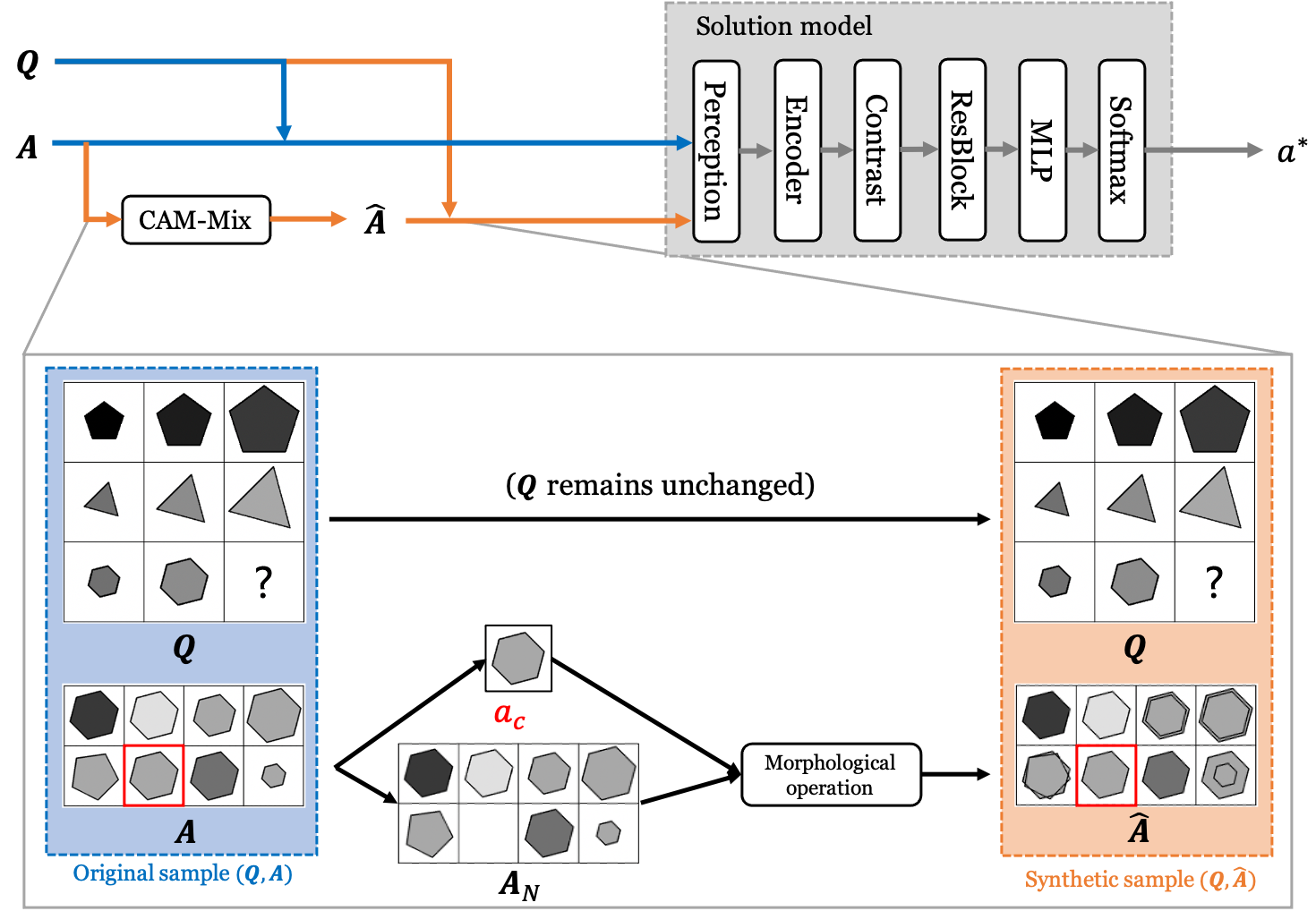}
    \caption{An example of applying CAM-Mix to $a_{c}$ and $\boldsymbol{A}_N$ on the training set. 
    The new negative candidate answers $\hat{\boldsymbol{A}}_N$ used in training are generated by mixing up the correct candidate answer with original negative candidate answers. 
    }
    \label{example_mixup}
\end{figure*}

As a data augmentation method for RPM problems, we aim to increase the training samples by mixing up the candidate answers, i.e. taking pixel-wise morphological operation $\mathcal{OR}$ and $\mathcal{AND}$ of negative candidates and the correct candidate. Formally, given the training samples $(\boldsymbol{Q},\boldsymbol{A})$, where $\boldsymbol{A}$ contains the correct candidate $a_{c}$ and negative candidates $\boldsymbol{A}_N$: $\boldsymbol{A} = \{a_{c}\} \cup \boldsymbol{A}_N $, the objective is to generate more negative candidate answers $\hat{\boldsymbol{A}}_{N}$ using the proposed gray-level morphological operations:
\begin{align}
    & \forall a_i \in \boldsymbol{A}_{N}, \quad \hat{a}_i = \mathcal{OR}(a_i,a_{c}) \in \hat{\boldsymbol{A}}_{N}, \label{eq:6} \\
    & \forall a_i \in \boldsymbol{A}_{N}, \quad \hat{a}_i = \mathcal{AND}(a_i,a_{c}) \in \hat{\boldsymbol{A}}_{N}. \label{eq:7}
\end{align}

\begin{figure}[pt]
    \centering
    \subfigure[(a) Distribution of original data]{
        \begin{minipage}[t]{0.45\linewidth}
            \centering
            \includegraphics[width=4cm]{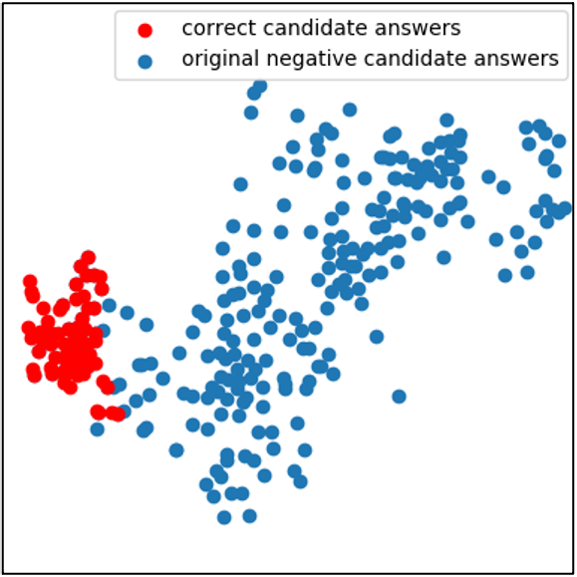}
        \end{minipage}
    }
    \subfigure[(b) Distribution of augmented data]{
        \begin{minipage}[t]{0.45\linewidth}
            \centering
            \includegraphics[width=4cm]{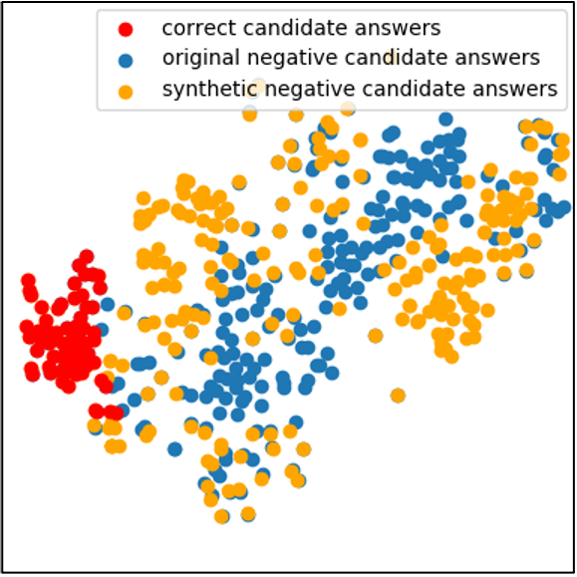}
        \end{minipage}
    }
    \caption{Visualization of applying the proposed CAM-Mix to a mini-batch of the RAVEN dataset. With more synthetic data falling onto uncharted space, a more precise decision boundary round the space spanned by the correct candidate answers could be obtained. The red dots represent correct candidates, the blue dots represent original negative candidates and the orange dots represent newly generated synthetic negative candidates. }
    \label{explanation}
\end{figure}

\begin{figure*}[pt]
    \centering
    \subfigure[(a) With vanilla mixup]{
        \begin{minipage}[t]{0.3\linewidth}
            \centering
            \includegraphics[width=4.3cm]{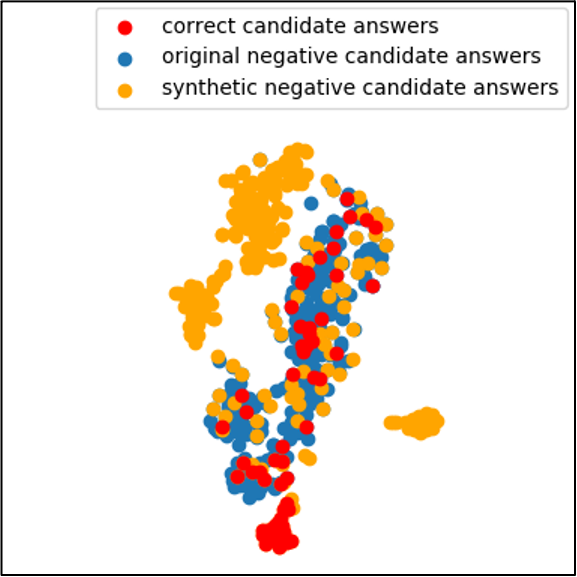}
        \end{minipage}
    }
    \subfigure[(b) With CAM-Mix ($\mathcal{OR}$)]{
        \begin{minipage}[t]{0.3\linewidth}
            \centering
            \includegraphics[width=4.3cm]{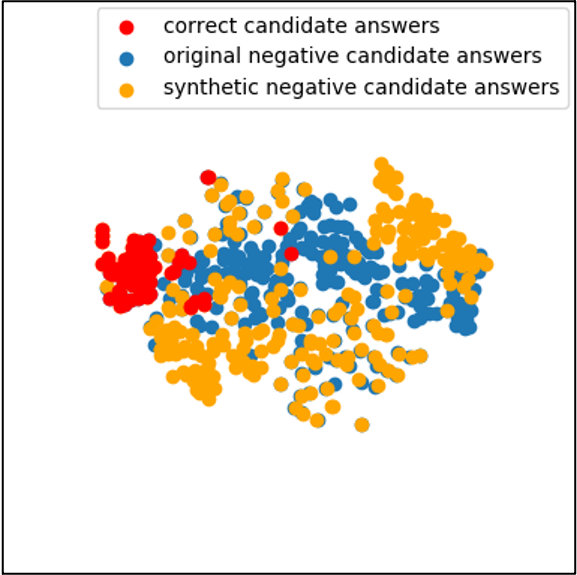}
        \end{minipage}
    }
    \subfigure[(c) With CAM-Mix ($\mathcal{AND}$)]{
        \begin{minipage}[t]{0.3\linewidth}
            \centering
            \includegraphics[width=4.3cm]{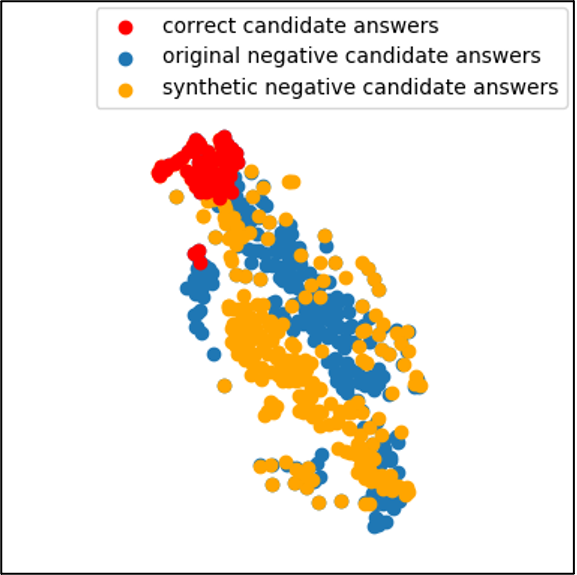}
        \end{minipage}
    }
    \caption{The visualization of different data augmentation methods applied to the solution model CoPINet \cite{zhang2019learning} on the RAVEN dataset \cite{zhang2019raven}, (a) the vanilla mixup \cite{zhang2018mixup} (b) the proposed CAM-Mix ($\mathcal{OR}$) (c) the proposed CAM-Mix ($\mathcal{AND}$). The figures are derived by extracting features from the last hidden layer of the solution models and projecting them into the first two principle dimensions for visualization. }
    \label{fig:visualization}
\end{figure*}

We build a set of extra training samples $(\boldsymbol{Q},\hat{\boldsymbol{A}})$ in Fig.~\ref{example_mixup}, where $\hat{\boldsymbol{A}} = \{a_{c}\} \cup \hat{\boldsymbol{A}}_{N}$. By mixing up the negative candidate answers with the correct one, those newly generated answers have partial information from the correct answer. As a result, the newly generated $\hat{\boldsymbol{A}}_{N}$ could define a more precise decision boundary around the space spanned by the correct answers, as evidenced later in Fig.~\ref{explanation} and Fig.~\ref{fig:visualization}.

The proposed data augmentation method generates extra synthetic training data by partially introducing the information of the correct answer into negative candidates. Such a formulation allows the models to explore possible interpolation functions on the `uncharted space' that was not covered by the original training data, as illustrated by Fig.~\ref{explanation}. By utilizing the proposed data augmentation scheme, extra negative candidate answers (orange dots) introduce extra decision support and a better estimation of the decision boundary of the model. In addition, the proposed method generates negative candidate answers using the original finite set of visual attribute values without explicitly extracting them, and the generated visual attribute values are still within the finite set. The proposed method hence does not significantly complicate the visual recognition problem, while the newly generated samples could provide additional decision support to the subsequent logical reasoning problem.

\section{Experimental results}
\label{sec:experiment}

We evaluate the proposed data augmentation method on two benchmark datasets: RAVEN \cite{zhang2019raven} and I-RAVEN \cite{hu2021stratified}. We show that by utilizing the proposed CAM-Mix augmentation, we achieve a significant and consistent performance improvement compared with state-of-the-art approaches. 

\subsection{Dataset description}

\begin{figure}[htbp]
    \centering
    \subfigure[{Center}]{
        \begin{minipage}[t]{0.05\textwidth}
            \centering
            \includegraphics[width=1\textwidth]{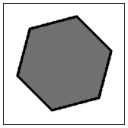}
        \end{minipage}
    }
    \subfigure[{2$\times$2Grid}]{
        \begin{minipage}[t]{0.05\textwidth}
            \centering
            \includegraphics[width=1\textwidth]{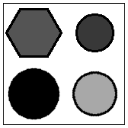}
        \end{minipage}
    }
    \subfigure[{3$\times$3Grid}]{
        \begin{minipage}[t]{0.05\textwidth}
            \centering
            \includegraphics[width=1\textwidth]{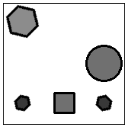}
        \end{minipage}
    }
    \subfigure[{L-R}]{
        \begin{minipage}[t]{0.05\textwidth}
            \centering
            \includegraphics[width=1\textwidth]{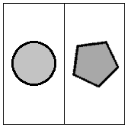}
        \end{minipage}
    }
    \subfigure[{U-D}]{
        \begin{minipage}[t]{0.05\textwidth}
            \centering
            \includegraphics[width=1\textwidth]{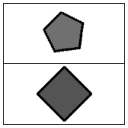}
        \end{minipage}
    }
    \subfigure[{O-IC}]{
        \begin{minipage}[t]{0.05\textwidth}
            \centering
            \includegraphics[width=1\textwidth]{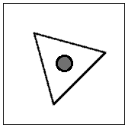}
        \end{minipage}
    }
    \subfigure[{O-IG}]{
        \begin{minipage}[t]{0.05\textwidth}
            \centering
            \includegraphics[width=1\textwidth]{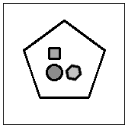}
        \end{minipage}
    }

    \caption{7 different configurations in the RAVEN \cite{zhang2019raven} dataset. The Center, Left-Right, Up-Down and O-IC configurations have determined Number and Position attributes. The remaining three, on the contrary, require to reason rules on the Number and Position attributes and hence are more complicated to solve. }
    \label{raven_config}
\end{figure}

\noindent \textbf{RAVEN} dataset \cite{zhang2019raven} has 70,000 samples.  Each question contains 16 images, including a 3$\times$3 image matrix with the last one missing, and 8 images of candidate answers. Given the first 8 images of the question, the objective is to deduce the reasoning rules from the first 8 images and choose the most suitable image from the 8 candidate answers as the correct one. The candidate answers are generated by permutation of visual attributes of the correct answer image, and each permuted image is derived by randomly shifting one attribute value. The dataset consists of 7 problem configurations and each sub-patch in a question matrix is constructed using 6 attributes (Angle, Number, Position, Type, Size, Color) and 4 underlying rules (Constant, Progressive, Arithmetic, Distribute\_three).
Examples of these configurations are shown in Fig.~\ref{raven_config}. 

\noindent \textbf{I-RAVEN} dataset \cite{hu2021stratified} is a variant of the RAVEN dataset \cite{zhang2019raven}. Hu \etal argue that there exist defects in choice panels of the original RAVEN dataset, which may lead to a shortcut to the correct answer, and hence a dataset with impartial candidate answers is built. 
The negative candidate answers of the I-RAVEN dataset are generated by hierarchically permuting one attribute of the correct answer in three iterations. In each iteration, two child nodes are generated, where one node remains the same with the parent node and the other permutes once. 
The I-RAVEN dataset shares the same settings and configurations with the original RAVEN dataset except the distinction in generated negative candidate answers. As shown later, the I-RAVEN dataset is more difficult than the original RAVEN dataset, after removing the shortcut.

\subsection{Experimental settings}
We apply the proposed data augmentation to the current best methods in the literature, i.e. CoPINet \cite{zhang2019learning} and SRAN \cite{hu2021stratified}, for RAVEN and I-RAVEN, respectively. Both models are compared to various recently published methods \cite{santoro2018measuring,zhang2019raven,zheng2019abstract}. 
We test and compare the performance of the vanilla mixup \cite{zhang2018mixup} and the proposed CAM-Mix ($\mathcal{OR}$, $\mathcal{AND}$) on both models. For the vanilla mixup, we follow the original formulation and hyperparameter settings in \cite{zhang2018mixup}, which alpha-blends the candidate images as well as the associated labels with $\lambda \sim Beta(\alpha,\alpha)$, $\alpha \in (0,\infty)$. 
The synthetic dataset is combined with the original dataset to train the two models. Specifically, both datasets originally consist of 42,000 samples for training, 14,000 for validation, and 14,000 for testing. With the proposed data augmentation method, we produce 42,000 extra synthetic samples, forming up to 84,000 training samples in total. Neither validation nor testing data are modified. 

\subsection{Visualization of data augmentation}

\begin{figure}[!htbp]
    \centering
    \includegraphics[width=1\linewidth]{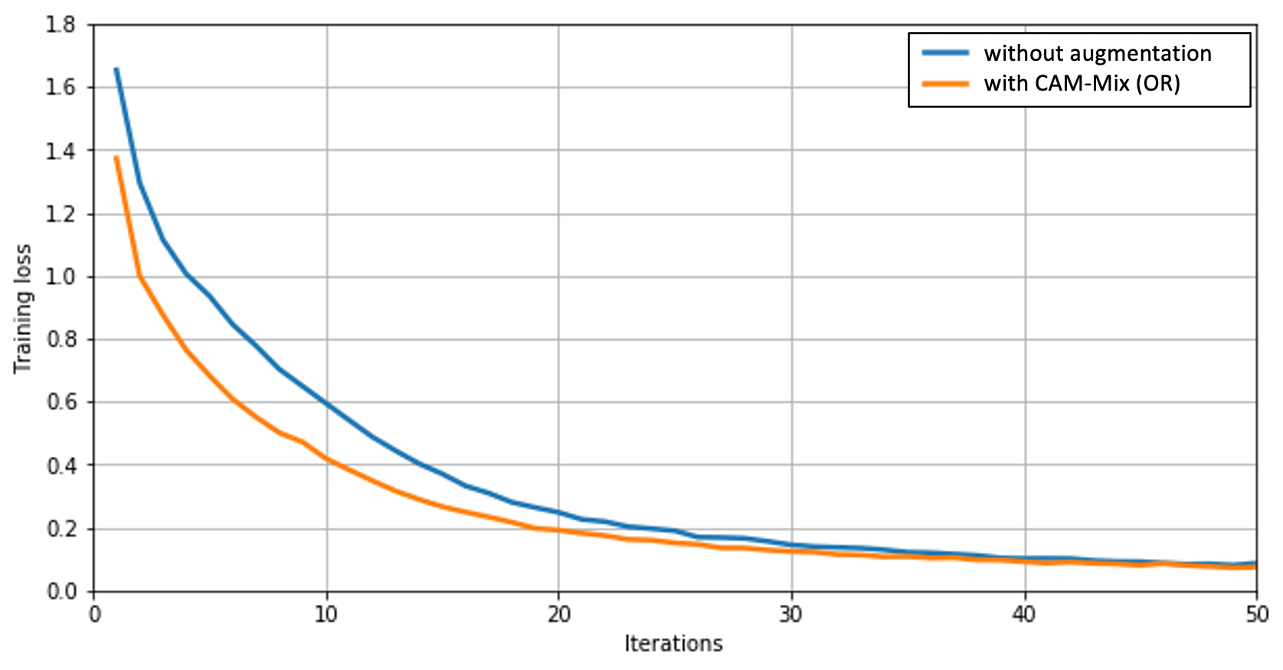}
    \caption{Training loss transitions when the proposed CAM-Mix data augmentation is and is not applied to train the SRAN model \cite{hu2021stratified} on the I-RAVEN dataset. Clearly, with the proposed data augmentation techniques, the training loss drops much faster than the case without the augmentation. This demonstrates the effectiveness of the proposed data augmentation in helping the model to converge faster. }
    \label{fig:loss_plot}
\end{figure}

In Fig.~\ref{fig:loss_plot}, we firstly demonstrate that the proposed CAM-Mix ($\mathcal{OR}$) data augmentation could benefit the model training process, as the training loss converges faster than the situation where the data augmentation is not applied.

Secondly, we show that the proposed CAM-Mix can help to define a clearer decision boundary around the cluster of correct answer samples with the new synthetic samples. For a better visualization, we plot mini-batches of normalized training data for three different data augmentation methods: vanilla mixup, proposed CAM-Mix ($\mathcal{OR}$) and CAM-Mix ($\mathcal{AND}$).
The features extracted from the last hidden layer of applied solution model CoPINet \cite{zhang2019learning} can be visualized as a 2-D plot as shown in Fig.~\ref{fig:visualization} after dimension reduction to obtain the first two principle components. We plot the original data and the synthetic data generated by the vanilla mixup \cite{zhang2018mixup}, the proposed CAM-Mix ($\mathcal{OR}$) and CAM-Mix ($\mathcal{AND}$), shown in three subplots of Fig.~\ref{fig:visualization} respectively. As the optimal model parameters of solution models for different sets of training data are different, the data distributions are shown in different subspaces and not directly comparable.

From Fig.~\ref{fig:visualization}(a), we can see that although the vanilla mixup can generate a certain amount of data falling into the uncharted space around the original candidate answers, the correct candidate answers cannot be well separated from either the original negative candidate answers or the synthetic negative candidate answers. The generated synthetic data do not help the model converge to generate a subspace where the correct candidate answers could be well separated from the negative candidate answers.

Contrastingly, the proposed CAM-Mix could not only generate a remarkable amount of data that fills in the uncharted space around the correct candidate answers, but also form up a clearer decision boundary around the space spanned by the correct candidate answers, as visualized in Fig.~\ref{fig:visualization}(b) and Fig.~\ref{fig:visualization}(c). This demonstrates the effectiveness of the newly-generated synthetic data to help the model converge to generate a subspace where positive answers and negative answers can be well separated. The synthetic data well fall into the uncharted space around the correct candidate answers. As a plug-and-play tool, the proposed CAM-Mix is advantageous to the visual reasoning systems and various RPM solution models. 

\subsection{Experimental results on the RAVEN dataset}

\begin{table*}[pt]
    \setlength{\belowcaptionskip}{5mm}
    \centering
    \small
    \caption{The reasoning accuracy on the RAVEN dataset. The results for other models are referred from \cite{zhang2019learning}. Overall, all three data augmentation methods could improve the classification performance of the state-of-the-art method, CoPINet. Among these methods, the proposed two methods outperform the vanilla mixup \cite{zhang2018mixup}, the original CoPiNet \cite{zhang2019learning} and other solution models.}
    \label{tab:raven}
    \begin{tabular}{l|c|ccccccc}
        \toprule
        Methods & Accuracy & Center & 2$\times$2Grid & 3$\times$3Grid & Left-Right & Up-Down & O-IC & O-IG  \\
        \midrule 
        WReN \cite{santoro2018measuring} (ICML, 2018) & 34.0\% & 58.4\% & 38.9\% & 37.7\% & 21.6\% & 19.7\% & 38.8\% & 22.6\% \\
        ResNet + DRT \cite{zhang2019raven} (CVPR, 2019) & 59.6\% & 58.1\% & 46.5\% & 50.4\% & 65.8\% & 67.1\% & 69.1\% & 60.1\% \\
        LEN \cite{zheng2019abstract} (NeurIPS, 2019) & 72.9\% & 80.2\% & 57.5\% & 62.1\% & 73.5\% & 81.2\% & 84.4\% & 71.5\% \\
        CoPINet \cite{zhang2019learning} (NeurIPS, 2019)  & 91.4\% & 95.1\% & 77.5\% & 78.9\% & 99.1\% & 99.7\% & 98.5\% & 91.4\% \\
        \midrule
        {CoPINet + vanilla mixup} & {92.0\%} & {95.9\%} & {78.2\%} & {80.7\%} & {99.1\%} & {99.6\%} & {98.5\%} & {92.2\%} \\
        {CoPINet + Proposed CAM-Mix ($\mathcal{OR}$)} & \textbf{93.3\%} & {97.8\%} & \textbf{79.2\%} & {82.2\%} & \textbf{99.9\%} & \textbf{99.8\%} & \textbf{99.6\%} & \textbf{94.6\%} \\
        {CoPINet + Proposed CAM-Mix ($\mathcal{AND}$)} & {93.2\%} & \textbf{98.5\%} & {79.0\%} & \textbf{83.5\%} & {99.7\%} & \textbf{99.8\%} & {99.1\%} & {93.0\%} \\
        \bottomrule
    \end{tabular}
\end{table*}

On the RAVEN dataset, Table~\ref{tab:raven} demonstrates an maximum improvement of 1.9\% on the mean accuracy by applying the proposed CAM-Mix ($\mathcal{OR}$) on top of the state-of-the-art model CoPINet \cite{zhang2019learning}, and consistent improvements can be witnessed on all the problem configurations. Particularly, on Center, 2$\times$2Grid, 3$\times$3Grid configurations, more than 2.7\% of performance improvement can be achieved compared with CoPINet. The proposed method also significantly outperforms other solution models on the RAVEN dataset.

Compared to traditional vanilla mixup \cite{zhang2018mixup}, the proposed CAM-Mix ($\mathcal{OR}$) achieves a performance gain of 1.3\% on the RAVEN dataset. Serving as a data augmentation technique, the vanilla mixup generates extra training samples that may assist the training process, and consequently gains an improvement of 0.6\% compared to CoPINet. However, it introduces new attribute values by pair-wise linear interpolation of intensity values as well as the associated labels. As a result, it can only have limited positive effect and it is far less effective compared with the proposed CAM-Mix, which truly mixes up the visual attributes of candidate answers, without explicitly extracting them or generating new attributes values. The proposed CAM-Mix can hence generate a better decision boundary around the space spanned by the correct samples, and hence perform better. 

\begin{figure}[!htbp]
    \centering
    \includegraphics[width=1\linewidth]{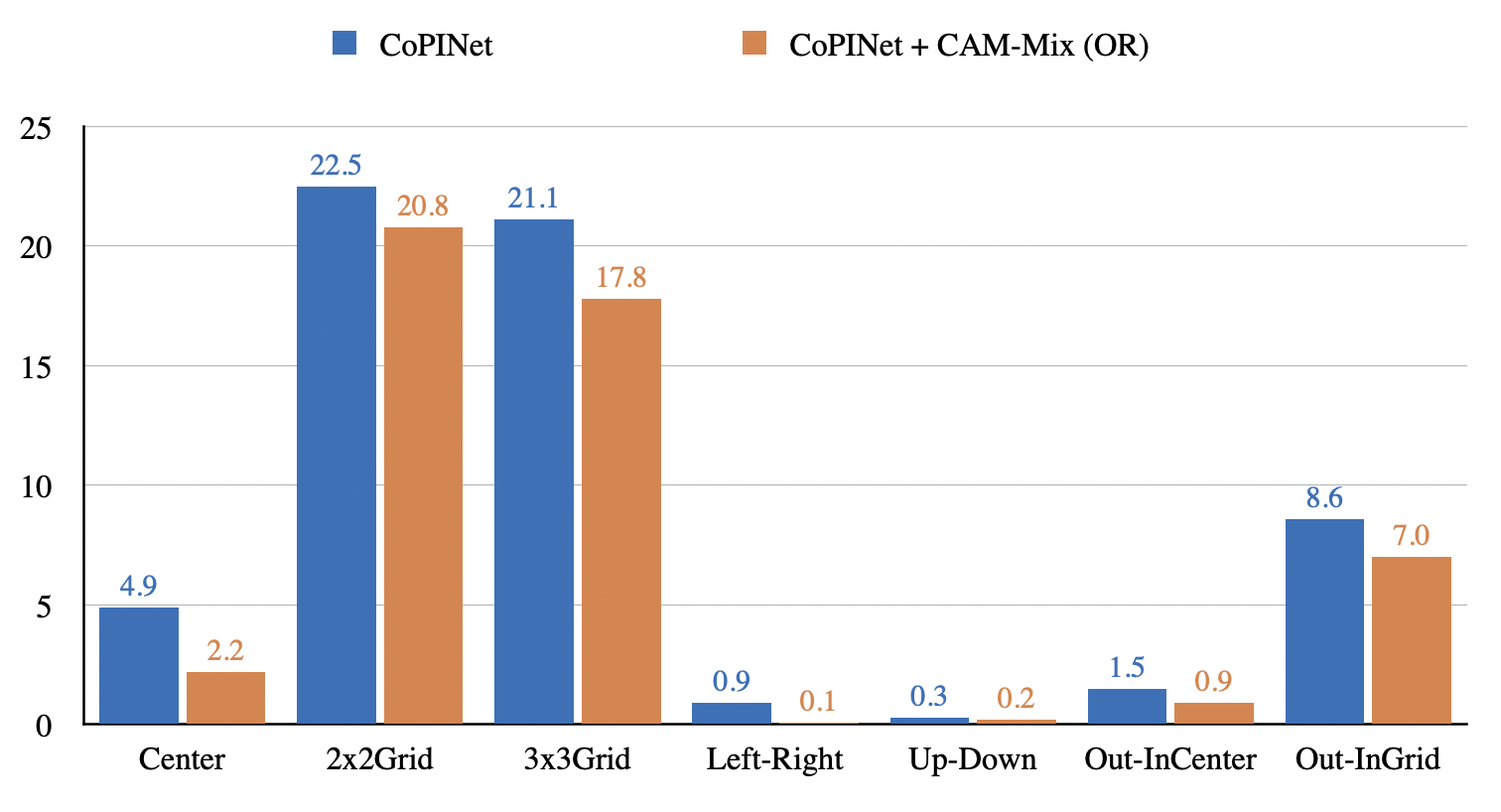}
    \caption{Error rate comparison of the reasoning results with and without proposed CAM-Mix ($\mathcal{OR}$) on the RAVEN dataset using the solution model CoPINet \cite{zhang2019learning}. After utilizing the proposed data augmentation method on the CoPINet \cite{zhang2019learning}, the error rates on all 7 configurations drop.}
    \label{fig:error_on_raven}
\end{figure}

We plot the error rate comparison in Fig.~\ref{fig:error_on_raven} with and without the proposed CAM-Mix ($\mathcal{OR}$), and consistent improvements is demonstrated on all 7 problem configurations. Notably, the proposed CAM-Mix can even strengthen the model reasoning capability on configurations that was almost perfectly resolved (e.g. Left-Right, Up-Down and Out-InCenter). 

\subsection{Experimental results on the I-RAVEN dataset}

\begin{table*}[h]
    \setlength{\belowcaptionskip}{5mm}
    \centering
    \small
    \caption{The reasoning accuracy on the I-RAVEN dataset. The results for other models are referred from \cite{hu2021stratified}. All three data augmentation methods significantly improve the accuracy of SRAN \cite{hu2021stratified}. Among these, the proposed CAM-Mix ($\mathcal{AND}$) significantly improves the mean accuracy by 12.9\%, and significantly outperforms the vanilla mixup by 9.3\%. }
    \label{tab:i-raven}
    \begin{tabular}{l|c|ccccccc}
        \toprule
        Methods & Accuracy & Center & 2$\times$2Grid & 3$\times$3Grid & Left-Right & Up-Down & O-IC & O-IG \\
        \midrule
        WReN \cite{santoro2018measuring} (ICML, 2018) & 23.8\% & 29.4\% & 26.8\% & 23.5\% & 21.9\% & 21.4\% & 22.5\% & 21.5\% \\
        ResNet+DRT \cite{zhang2019raven} (CVPR, 2019) & 40.4\% & 46.5\% & 28.8\% & 27.3\% & 50.1\% & 49.8\% & 46.0\% & 34.2\% \\
        LEN \cite{zheng2019abstract} (NeurIPS, 2019) & 41.4\% & 56.4\% & 31.7\% & 29.7\% & 44.2\% & 44.2\% & 52.1\% & 31.7\% \\
        CoPINet \cite{zhang2019learning} (NeurIPS, 2019) & 46.1\% & 54.4\% & 36.8\% & 31.9\% & 51.9\% & 52.5\% & 52.2\% & 42.8\% \\
        SRAN \cite{hu2021stratified} (AAAI, 2021) & 60.8\% & 78.2\% & 50.1\% & 42.4\% & 70.1\% & 70.3\% & 68.2\% & 46.3\% \\
        \midrule
        {SRAN + vanilla mixup} & {64.4\%} & {85.0\%} & {49.1\%} & {40.8\%} & {75.6\%} & {74.5\%} & {77.4\%} & {48.2\%} \\
        {SRAN + Proposed CAM-Mix ($\mathcal{OR}$)} & {73.0\%} & {93.3\%} & {55.6\%} & {51.8\%} & {82.8\%} & {82.3\%} & {88.1\%} & \textbf{57.3\%} \\
        {SRAN + Proposed CAM-Mix ($\mathcal{AND}$)} & \textbf{73.7\%} & \textbf{93.9\%} & \textbf{57.7\%} & \textbf{52.5\%} & \textbf{84.0\%} & \textbf{84.5\%} & \textbf{88.9\%} & {54.4\%} \\
        \bottomrule
    \end{tabular}
\end{table*}

\begin{table*}[pb]
    \setlength{\belowcaptionskip}{5mm}
    \centering
    \small
    \footnotesize
    \caption{Comparison for different input image sizes on the I-RAVEN dataset. The larger the image size, the better the classification performance. On both datasets, the proposed CAM-Mix method consistently and significantly boosts the performance of the best performed method, SRAN \cite{hu2021stratified}, on the I-RAVEN dataset. }
    \label{tab:sanity}
    \subtable[80$\times$80]{
    \begin{tabular}{l|c|ccccccc}
        \toprule
        Methods & Accuracy & Center & 2$\times$2Grid & 3$\times$3Grid & Left-Right & Up-Down & O-IC & O-IG \\
        \midrule
        CoPINet \cite{zhang2019learning} & 46.1\% & 54.4\% & 36.8\% & 31.9\% & 51.9\% & 52.5\% & 52.2\% & 42.8\% \\
        SRAN \cite{hu2021stratified} & 46.8\% & 61.1\% & 34.3\% & 30.9\% & 52.2\% & 50.9\% & 56.4\% & 41.9\% \\
        \midrule
        {SRAN + Proposed CAM-Mix ($\mathcal{OR}$)} & {60.6\%} & \textbf{86.6\%} & \textbf{46.5\%} & 41.3\% & \textbf{62.7\%} & 61.0\% & \textbf{77.7\%} & 47.7\% \\
        {SRAN + Proposed CAM-Mix ($\mathcal{AND}$)} & \textbf{60.9\%} & 85.6\% & 46.4\% & \textbf{42.1\%} & 62.5\% & \textbf{62.5\%} & 76.7\% & \textbf{50.2\%} \\
        \bottomrule
    \end{tabular}
    }
    \qquad
    \subtable[224$\times$224]{
    \begin{tabular}{l|c|ccccccc}
        \toprule
        Methods & Accuracy & Center & 2$\times$2Grid & 3$\times$3Grid & Left-Right & Up-Down & O-IC & O-IG \\
        \midrule
        CoPINet \cite{zhang2019learning} & 50.9\% & 57.2\% & 37.3\% & 32.8\% & 61.5\% & 62.5\% & 60.5\% & 44.3\% \\
        SRAN \cite{hu2021stratified} & 60.8\% & 78.2\% & 50.1\% & 42.4\% & 70.1\% & 70.3\% & 68.2\% & 46.3\% \\
        \midrule
        {SRAN + Proposed CAM-Mix ($\mathcal{OR}$)} & {73.0\%} & {93.3\%} & {55.6\%} & {51.8\%} & {82.8\%} & {82.3\%} & {88.1\%} & \textbf{57.3\%} \\
        {SRAN + Proposed CAM-Mix ($\mathcal{AND}$)} & \textbf{73.7\%} & \textbf{93.9\%} & \textbf{57.7\%} & \textbf{52.5\%} & \textbf{84.0\%} & \textbf{84.5\%} & \textbf{88.9\%} & {54.4\%} \\
        \bottomrule
    \end{tabular}
    }
\end{table*}

The I-RAVEN dataset has balanced candidate answer sets and cannot be resolved by building Contrast Module $\mathcal{C}$ on candidate answers, and hence is more complicated for existing solution models as shown in experimental results in Table~\ref{tab:i-raven}. The proposed data augmentation method is applied on top of the state-of-the-art model, SRAN \cite{hu2021stratified}, and gains a remarkable improvement on the I-RAVEN dataset, i.e. the proposed CAM-Mix ($\mathcal{AND}$) results in a performance gain of 12.9\% on the average accuracy. On all 7 different configurations, the proposed data augmentation method could significantly boost the performance of the previous best model, SRAN \cite{hu2021stratified}, on the I-RAVEN dataset. The largest performance gain on the Out-InCenter configuration is more than 20\%. 

\begin{figure}[!htbp]
    \centering
    \includegraphics[width=1\linewidth]{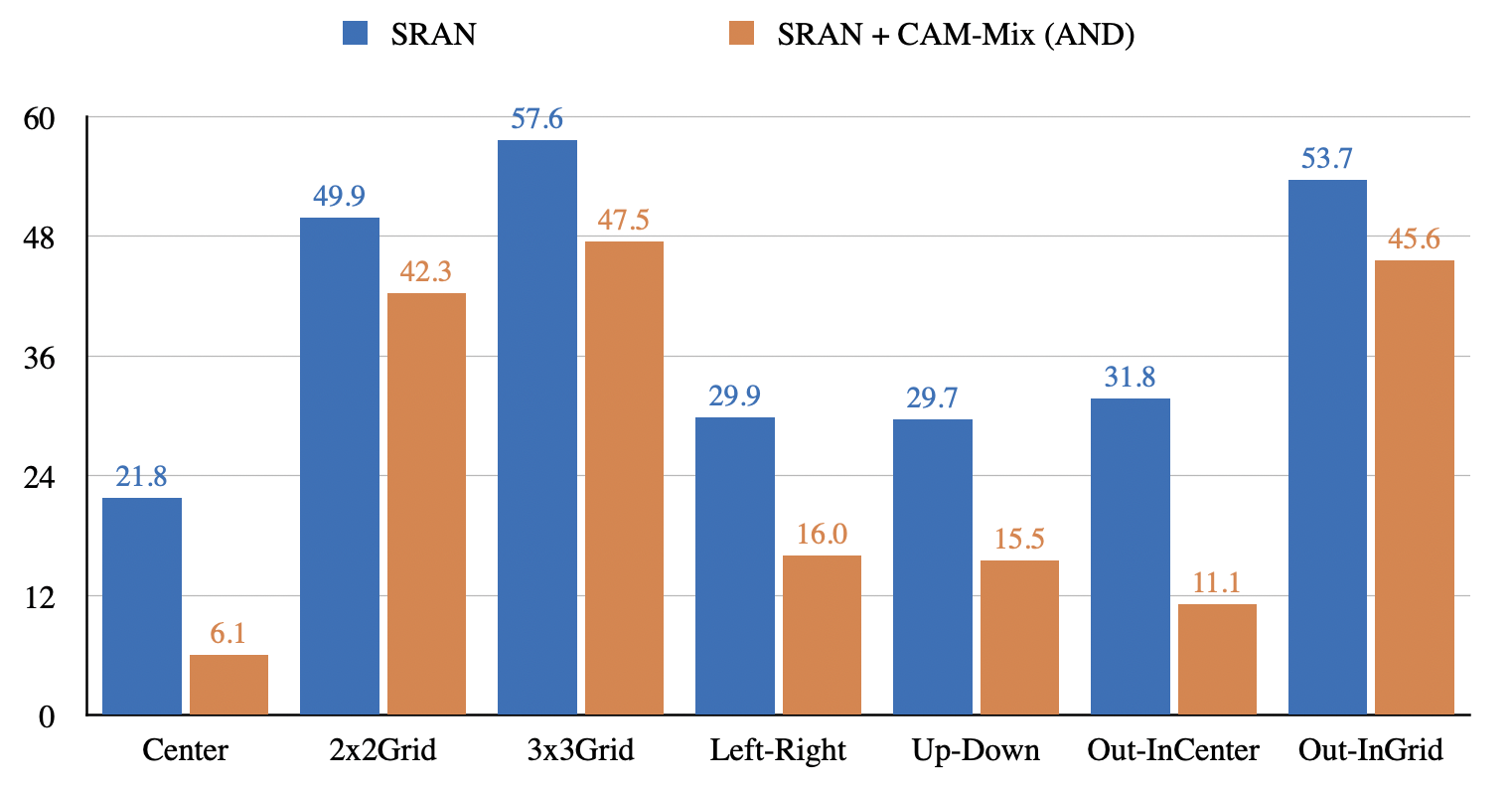}
    \caption{Error rate comparison of the reasoning results with and without proposed CAM-Mix ($\mathcal{AND}$) on the I-RAVEN dataset using the solution model SRAN \cite{hu2021stratified}. On all 7 configurations, the proposed CAM-Mix ($\mathcal{AND}$) significantly reduces the error rates of the state-of-the-art solution model, SRAN \cite{hu2021stratified}.}
    \label{fig:error_on_iraven}
\end{figure}

The proposed CAM-Mix ($\mathcal{AND}$) achieves a performance gain of 9.3\% on the RAVEN dataset compared to traditional vanilla mixup \cite{zhang2018mixup}. The results further show the limited capability that vanilla mixup brings to existing solution models compared to the proposed CAM-Mix. We plot the error rate comparison in Fig.~\ref{fig:error_on_iraven} with and without the proposed CAM-Mix. By utilizing the proposed CAM-Mix ($\mathcal{AND}$) on top of the state-of-the-art solution model, SRAN \cite{hu2021stratified}, the error rates on all 7 configurations are significantly reduced. As a generalized plug-and-play tool, the proposed method could boost the performance of various state-of-the-art models, as evidenced in Table \ref{tab:raven} and \ref{tab:i-raven}. 

\subsection{The impact of image size}

The results reported in \cite{hu2021stratified} are based on the images of size 224$\times$224, but other models \cite{santoro2018measuring,zhang2019raven,zhang2019learning,zheng2019abstract} are evaluated on the images of size 80$\times$80. For a fair comparison, we conduct a series of experimental studies on the I-RAVEN dataset with images of these two different sizes. We also evaluate the performance of CoPINet model on the I-RAVEN dataset with image size of 224$\times$224. The mean accuracy arises from 46.1\% to 50.9\%, which shows the positive effect of a larger image size. SRAN outperforms CoPINet on the I-RAVEN dataset by 9.9\% using the image size of 224$\times$224. The accuracy is further boosted to 60.6\% and 73.0\% by utilizing the proposed CAM-Mix ($\mathcal{OR}$), and 60.9\% and 73.7\% by utilizing the proposed CAM-Mix ($\mathcal{AND}$) on the I-RAVEN dataset with image size of 80$\times$80 and 224$\times$224, respectively. All the experimental results demonstrate the effectiveness of the proposed CAM-Mix method.

\section{Conclusion}
\label{sec:conclusion}
In this paper, we propose a data augmentation method, CAM-Mix, for RPM problems by generating more negative candidate answers using gray-level morphological operations. The proposed method mixes-up the visual attributes of original images, without explicitly extracting them. The generated negative candidate answers are semantically similar to the correct answers, and hence a better decision boundary could be defined around the space spanned by the correct answers. The proposed approach is systematically evaluated on benchmark datasets and outperforms the state-of-the-art models. CAM-Mix can serve as a plug-and-play tool to existing solution models and receive a consistent accuracy improvement. 
Additionally, the proposed data augmentation method is potentially widely beneficial in various multiple-choice problems. In future, we will explore the feasibility of applying the proposed approach on other problems.

\section*{Declaration of Competing Interest}
The authors declare that they have no known competing financial interests or personal relationships that could have appeared to influence the work reported in this paper. 

\section*{Acknowledgment}
This research is supported in part by the National Natural Science Foundation of China under Grant 72071116, and in part by the Ningbo Municipal Bureau Science and Technology under Grants 2019B10026 and 2017D10034.

\bibliography{mybib}

\end{document}